\documentclass[10pt,twocolumn,letterpaper]{article}

\usepackage{wacv}
\usepackage{times}
\usepackage{epsfig}
\usepackage{graphicx}
\usepackage{amsmath}
\usepackage{amssymb}
\usepackage{booktabs}

\usepackage[ruled, vlined]{algorithm2e}
\usepackage{xcolor}
\usepackage{multirow}
\usepackage[accsupp]{axessibility}

\SetArgSty{textnormal}
\SetKwInOut{Input}{Input}
\SetKwInOut{Output}{Output}
\SetKwIF{If}{ElseIf}{Else}{if}{then}{else if}{else}{endif}
\SetKwFor{For}{for}{do}{endfor}

\def\wacvPaperID{627} 

\wacvalgorithmstrack   

\wacvfinalcopy 

\ifwacvfinal
\usepackage[breaklinks=true,bookmarks=false]{hyperref}
\else
\usepackage[pagebackref=true,breaklinks=true,colorlinks,bookmarks=false]{hyperref}
\fi

\begin{document}

\title{Weakly Supervised Face Naming with Symmetry-Enhanced Contrastive Loss}


\author{Tingyu Qu\textsuperscript{1}\footnotemark[1], Tinne Tuytelaars\textsuperscript{2}, Marie-Francine Moens\textsuperscript{1}\\
\textsuperscript{1} Department of Computer Science, KU Leuven\\
\textsuperscript{2} Department of Electrical Engineering, KU Leuven\\
{\tt\small \{tingyu.qu, tinne.tuytelaars, sien.moens\}@kuleuven.be}
}

\maketitle
\thispagestyle{empty}

\begin{abstract}
We revisit the weakly supervised cross-modal face-name alignment task; that is, given an image and a caption, we label the faces in the image with the names occurring in the caption. Whereas past approaches have learned the latent alignment between names and faces by uncertainty reasoning over a set of images and their respective captions, in this paper, we rely on appropriate loss functions to learn the alignments in a neural network setting and propose \textbf{SECLA} and \textbf{SECLA-B}. \textbf{SECLA} is a \textbf{S}ymmetry-\textbf{E}nhanced \textbf{C}ontrastive \textbf{L}earning-based \textbf{A}lignment model that can effectively maximize the similarity scores between corresponding faces and names in a weakly supervised fashion. A variation of the model, SECLA-B, learns to align names and faces as humans do, that is, learning from easy to hard cases to further increase the performance of SECLA. More specifically, SECLA-B applies a two-stage learning framework: (1) Training the model on an easy subset with a few names and faces in each image-caption pair. (2) Leveraging the known pairs of names and faces from the easy cases using a bootstrapping strategy with additional loss to prevent forgetting and learning new alignments at the same time. We achieve state-of-the-art results for both the augmented Labeled Faces in the Wild dataset and the Celebrity Together dataset. In addition, we believe that our methods can be adapted to other multimodal news understanding tasks.
\end{abstract}

\renewcommand{\thefootnote}{\fnsymbol{footnote}}
\footnotetext[1]{We acknowledge the support from China Scholarship Council, CELSA/19/018, and MACCHINA project (KU Leuven C14/18/065).}

\section{Introduction}

With the rapid development of online media, news in multimedia format (e.g., text, image, and video) has become mainstream, which has driven the need for multimodal news understanding.

In multimodal news presentations, images and their respective captions enrich the content of news article text. Unlike in the conventional image captioning task \cite{conf/eccv/FarhadiHSYRHF10, conf/cvpr/VinyalsTBE15, conf/icml/XuBKCCSZB15}, where the contents of the image and its caption are often semantically well aligned, a news image-caption pair usually has weak semantic alignment, which makes the automated understanding of news images a challenging task.

\begin{figure}[t]
  \centering
\includegraphics[width=5cm]{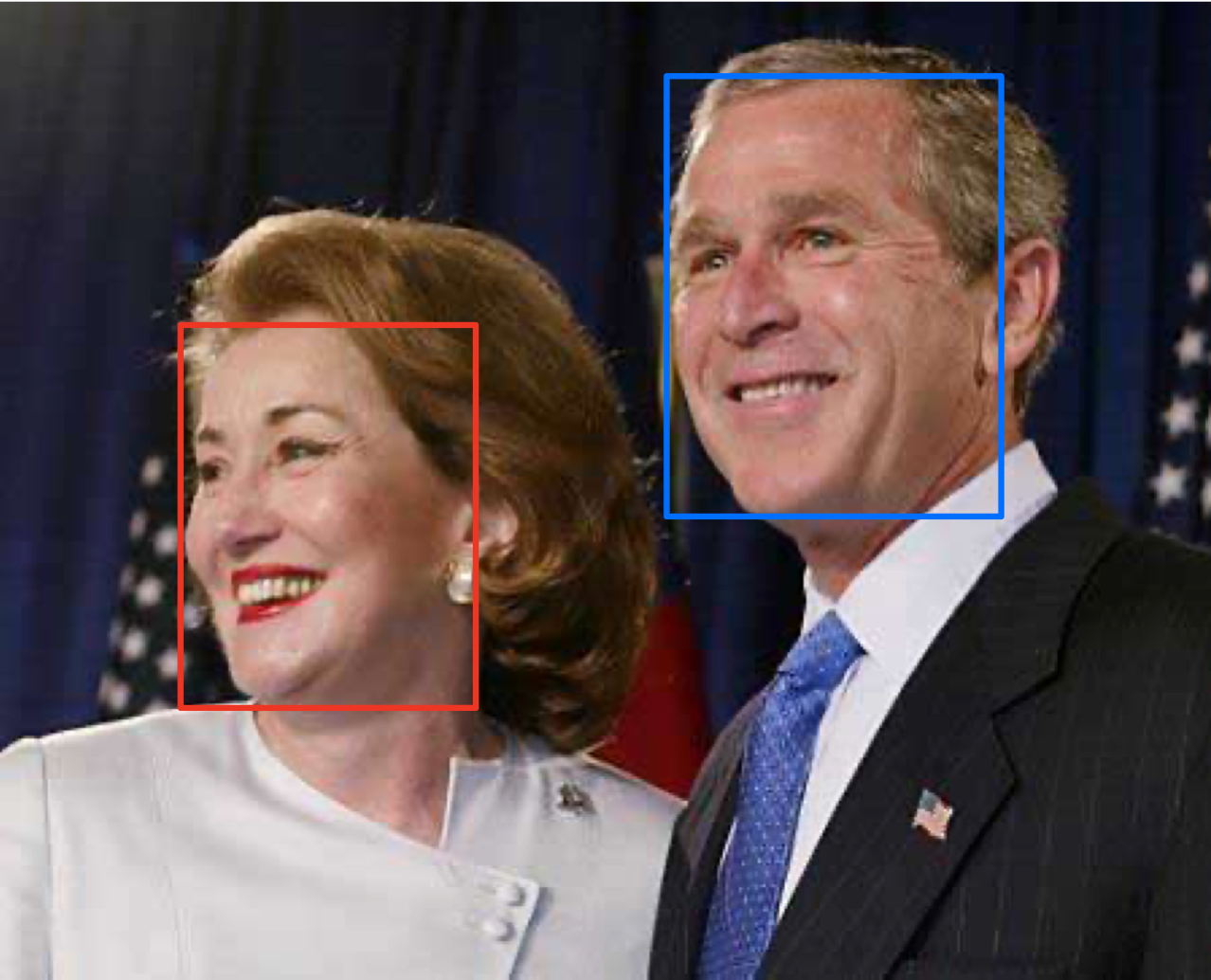}
\caption{News Image with Caption: President \textcolor{blue}{\textbf{George W. Bush}} attends a fundraiser for former Secretary of Transportation \textcolor{red}{\textbf{Elizabeth Dole}} while in Greensboro, North Carolina, July 25, 2002. Dole is running for the U.S. Senate seat to be vacated in November by Senator \textcolor{orange}{\textbf{Jesse Helms}}.}
\label{fig:example_lfw}
\end{figure}

Frequently, nonfamous people are shown in news images, for which no separate face recognizer is trained. However, in such cases, weak supervision of the caption helps in assigning names to faces. Finding which name in the caption corresponds to which face in the image is referred to as the face-name alignment task. Interpreting the image and the caption 
in Figure \ref{fig:example_lfw}, 
we will assign the face in the blue bounding box to the name George W. Bush, the face in the red box to the name Elizabeth Dole and 
"NOFACE" (i.e., names without corresponding faces) will be assigned to the name "Jesse Helms".

In this work, we revisit the weakly supervised face-name alignment task. The task is tailored to real-world multimodal news understanding by disambiguating names and faces of different persons in a weakly supervised setting, where explicit annotations of faces are lacking. This task is different from face recognition tasks \cite{schroff2015facenet}, which usually require carefully curated labels for training. With the significant number of news articles published every day, it is almost impossible to maintain such a database for face recognition.

Specifically, to address the weakly supervised face-name alignment problem, we propose \textbf{SECLA}, a \textbf{S}ymmetry-\textbf{E}nhanced \textbf{C}ontrastive \textbf{L}earning \textbf{A}lignment model consisting of pretrained feature extractors for feature extraction and multilayer perceptron (MLP) projectors for mapping features of faces and names onto the same embedding space.
Given projected features, SECLA is trained with symmetry-enhanced loss functions, including a bidirectional contrastive loss function and an agreement loss function.
The design of the symmetry-enhanced loss function is motivated by the assumption that each face corresponds to at least one name (i.e., dense alignment), which is often violated when null links are present, that is, names with no corresponding faces or faces with no corresponding names, as pointed out by \cite{hessel-etal-2019-unsupervised, conf/nips/KarpathyJL14}.

SECLA achieves better performance than the previous state-of-the-art (SOTA) model \cite{5332299}. Moreover, for easier subsets that contain few faces and names in each image-caption pair, SECLA can achieve (near-)perfect alignment results, which motivates us to propose
\textbf{SECLA-B}, a model that learns with a two-stage learning strategy and bootstrapping. We first train a SECLA model on an easy subset and then add the rest of the data into the second stage of training with loss functions that can utilize the known pairs of faces and names and their prototypes.
SECLA-B further improves the SECLA performance.
Thanks to the simple architectures, our proposed methods have great potential to be adapted to other multimodal news understanding tasks.

In summary, our contributions are threefold.
(1) We propose a contrastive learning-based model, called SECLA, that can accurately learn the alignment between faces and names in image caption pairs in a weakly supervised fashion.
(2) To address violations of the dense alignment assumption, we propose a novel symmetry-enhanced contrastive loss function that learns to disambiguate when the model is unsure about the correct alignment.
The SECLA model with symmetry-enhanced losses achieves SOTA performance for the task. We achieve (near-)perfect performance for the easy dataset with few faces and names in each image-caption pair.
(3) We propose a two-stage learning strategy with bootstrapping to utilize the (near-)perfect performance of SECLA on the easy subset. The model trained with this strategy, called SECLA-B, further improves the performance of SECLA.

\section{Related Works}

\begin{figure*}[t]
\begin{center}
\includegraphics[width=0.98\textwidth]{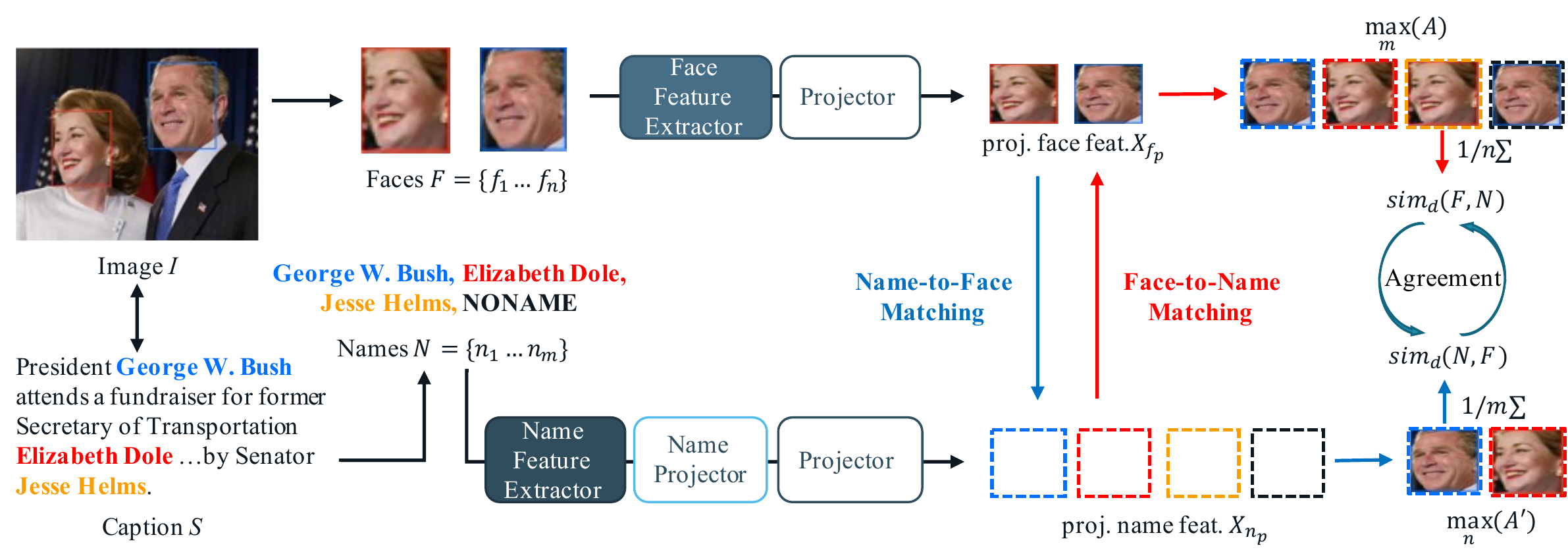}
\end{center}
\caption{The architecture of the proposed SECLA model. We first use feature extractors (FaceNet for faces and BERT for names) to obtain the features of faces $f_1,\ldots f_n$ in image I and names $n_1, \ldots, n_m$ (with one NONAME added) in caption $S$. The features are passed through projectors to a common embedding space. Then, we learn the alignment of faces and names in both directions, that is, the name-to-face and face-to-name alignments, with a contrastive loss. To further enhance the symmetry from the two directions, we apply an agreement loss to the similarity matrices that result from the computation of dense similarity measures. }
\label{fig:model}
\end{figure*}

Past studies on face-name alignment can be divided into two categories:

\textbf{Without external data.} Most face-name alignment studies have focused on exploring the relationship of image-caption pairs in a dataset without the use of external data mined from the web. These methods are typically built with an uncertainty reasoning framework.
Berg \etal~\cite{NIPS2004_03fa2f75, 1315253} proposed methods based on modified k-means clustering for captioned news images.
Guillaumin \etal~\cite{conf/cvpr/GuillauminMVS08} adopted a graph-based approach for retrieving all faces of the same person and aligning names to faces in each image-caption pair.
Su \etal~\cite{journals/mta/SuPFW16, journals/mta/SuPFWFC14} focused on an easier problem where only people with names occurring more than a certain number of times were studied and achieved satisfactory results.
Pham \etal~\cite{5332299} extended \cite{NIPS2004_03fa2f75} with a probabilistic approach under the 
EM
 framework. Given a good initialization of clusters with image-caption pairs, their method can make alignments in both directions, namely, name-to-face and face-to-name, using uncertainty reasoning. Their method achieved SOTA results on the Labeled Faces in the Wild (LFW) dataset. In this paper, we study the weakly supervised face-name alignment problem, as in \cite{5332299}.
Different from past works without external data, we do not rely on uncertainty reasoning; instead, we learn alignments in a neural network setting by designing appropriate loss functions that help disambiguate possible alignment issues in a weakly supervised setting.

\textbf{With external data.} Another line of research has focused on designing systems that utilize external data (mostly web data) to augment internal data, especially noncelebrities data that occur only a few times.
Zhang \etal~\cite{conf/ijcai/ZhangWLJXF13} explored the syntactic structure of captions for initializing the ranking of names and applied web mining to acquire extra data for rare celebrities.
Chen \etal~\cite{conf/mir/ChenFNJH15, journals/mms/ChenZDXG19} designed graph-based alignment methods to satisfy predefined constraints and used a Google image search for external data.
Recent work by Tian \etal~\cite{8880535} proposed a system that integrates deep multimodal analysis, cross-modal correlation learning, and web mining. Their proposed system is similar to that in \cite{conf/ijcai/ZhangWLJXF13}.

Apart from the face-name alignment task, our work is also closely related to a similar task, that is, \textbf{image-sentence matching}. Karpathy \etal~\cite{conf/nips/KarpathyJL14} discussed the violation of the dense alignment assumption and proposed an alignment model based on hinge loss. Hessel \etal~\cite{hessel-etal-2019-unsupervised} studied an unsupervised image-sentence matching problem. They proposed a dense similarity measure for calculating the similarity between image-sentence pairs and a hinge loss function with a top-k selection process to mitigate the violation of the dense alignment assumption \cite{hessel-etal-2019-unsupervised, conf/nips/KarpathyJL14}. We adopt a similar dense similarity measure, as detailed in Sec. \ref{sec:learning}.

Finally, our work is related to \textbf{visual grounding}.
Wang \etal~\cite{conf/emnlp/WangTSMY20} adopted a contrastive loss to learn the alignment between objects and phrases in a weakly supervised fashion. Cui \etal~\cite{Cui_2021_ICCV} proposed the human-centric visual grounding task called Who's Waldo, where the aim is to link anonymized names in the caption to the bounding boxes of people in the image. The authors added a NONAME constant to deal with null links.

\section{Methodology}

In this section, we formally describe the face-name alignment problem and introduce the proposed SECLA model. We then present the framework for training SECLA and a detailed explanation of two-stage learning with a bootstrapping strategy for SECLA-B. We conclude the section with the alignment scheme for the problem.

\subsection{Problem Formulation}

Given a news image $I$ with faces $F=\{ f_{1}, f_{2}, ..., f_{n} \}$ and the corresponding caption $S$ with names $N = \{ n_{1}, n_{2}, ..., n_{m} \}$, the goal of the face-name alignment task is to find the links between corresponding face $f_{i}$ and name $n_{j}$. We observe two types of links:
(1) Normal links: these are links with correctly detected faces and names, which appear in images and the corresponding captions.
(2) Null links: these are links containing faces without names ("NONAME") or names without faces ("NOFACE").

\subsection{Model Architecture}

As shown in Figure \ref{fig:model}, our SECLA model consists of two modules: (1) a feature extractor module for extracting features from face $f_{i}$ and name $n_{j}$ and (2) a projector module for projecting features onto a common embedding space.

The feature extractor module includes (1) a pretrained FaceNet \cite{schroff2015facenet} to extract embedding $\mathrm{X}^{i}_{f} \in \mathbb{R}^{d_{f}}$ for face $f_i$ and (2) a pretrained BERT \cite{devlin-etal-2019-bert} to produce name embedding $\mathrm{X}^{j}_{n} \in \mathbb{R}^{d_{n}}$ for name $n_j$.


The projector module consists of two types of projectors, (1) name projectors and (2) common projectors.
We first pass the name embedding $\mathrm{X}_{n}$ through a name projector $g_{n}(\cdot): \mathbb{R}^{d_{n}} \rightarrow  \mathbb{R}^{d_{f}}$ to obtain the projected embedding $\mathrm{X}_{n'}$, which has the same dimension as face embedding $\mathrm{X}_{f}$.
Then, we use two common projectors $g_{c}(\cdot): \mathbb{R}^{d_{f}} \rightarrow  \mathbb{R}^{d_{p}}$ to project $\mathrm{X}_{n'}$ and $\mathrm{X}_{f}$ onto the same embedding space with a lower dimension $d_{p}$.
We denote $\mathrm{X}_{n_{p}} \in \mathbb{R}^{d_{p}}$ and $\mathrm{X}_{f_{p}} \in \mathbb{R}^{d_{p}}$ as the projected features for $\mathrm{X}_{n'}$ and $\mathrm{X}_{f}$, respectively.
We calculate the similarity score between $f_i$ and $n_j$ as $sim(f_{i}, n_{j}) = (\mathrm{X}_{f_{p}}^i)^\mathrm{T} \cdot \mathrm{X}_{n_{p}}^j$ for model training and subsequent inferencing.

\subsection{Training the SECLA model}
\label{sec:learning}

To train our SECLA model, we propose the following loss functions that operate on projected features $\mathrm{X}_{n_{p}}^{j} $ and $\mathrm{X}_{f_{p}}^{i}$: (1) contrastive loss that maximizes the dense similarities between corresponding images and captions and (2) agreement loss that forces the symmetry from the similarity matrices computed from two directions.

\textbf{Contrastive Loss}. Inspired by previous works in caption ranking \cite{conf/cvpr/FangGISDDGHMPZZ15} and visual grounding \cite{conf/emnlp/WangTSMY20}, we apply the following contrastive loss computed from the face-to-name direction:

\begin{equation}
\label{eq:contras_fn}
\mathcal{L}_{f,n} = -log \frac{e^{sim_d(F,N)}}{\sum_{F_k \in batch} e^{sim_d(F_k ,N)} }
\end{equation}

The dense similarity measure $sim_d(F,N)$ between image $I$ and caption $S$ is defined as:

\begin{equation}
\label{eq:sim}
sim_d(F, N) = \frac{1}{n} \sum_{i=1}^{n} \max_{j} A_{i,j}
\end{equation}
where $A_{i,j} = sim( f_{i}, n_{j})$, $i=1,2,\ldots, n$, $j=1,2,\ldots, m$

Similar to the face-to-name contrastive loss $\mathcal{L}_{f,n}$, we define name-to-face contrastive loss as:

\begin{equation}
\label{eq:contras_nf}
\mathcal{L}_{n,f} = -log \frac{e^{sim_d(N,F)}}{\sum_{N_k \in batch} e^{sim_d(N_k ,F)} }
\end{equation}
with $sim_d(N, F) = \frac{1}{m} \sum_{j=1}^{m} \max_{i} A'_{j,i}$, where  $A'_{j,i} = sim(n_{j}, f_{i})$, $j=1,2,\ldots, m$, $i=1,2,\ldots, n$, .

In a weakly supervised setting, the connection between faces and names within the same image-caption pair remains unknown during training, and we must rely on dense similarity measures to learn a meaningful embedding space. However, as illustrated in \cite{hessel-etal-2019-unsupervised, conf/nips/KarpathyJL14}, the dense alignment assumption can be violated when null links exist. Unlike the phrase-region matching process in visual grounding, null links are extremely common in our task. As long as the numbers of faces and names in one image-caption pair are not equal or there exist noncelebrity faces, null links can be expected. Even with only one face in an image, we can have no corresponding name in the caption, and vice versa. To this end, unlike in visual grounding \cite{conf/emnlp/WangTSMY20}, where the impact of the more symmetrical contrastive loss function is limited, the symmetrical loss functions in our task are important. Consequently, we compute the contrastive loss functions in two directions, that is, face-to-name and name-to-face, to symmetrize the computation of the similarity scores.

\textbf{Agreement Loss}. The bidirectional contrastive loss can mitigate violations of the dense alignment assumption. However, this is not sufficient for our task. Especially in the early stage of training, with randomly initialized projectors, it is difficult for the model to differentiate between names or to guide the gradient flow in the right direction.

To further enhance the symmetry of our bidirectional model, we design an agreement loss for additional regularization during training, which has been proven effective in word alignment models \cite{chen-etal-2021-mask, liang-etal-2006-alignment}. Our agreement loss forces the dense similarity scores computed from
the face-to-name and name-to-face directions to be as close as possible.
Without it, we could have more aligned face-name pairs differ from two directions due to the $\max(\cdot)$ operator.
Specifically, for a mini batch of B samples, each containing \textit{m} names and \textit{n} faces, we obtain a tensor $Z$ of size $B\times B \times n \times m$ to represent all combinations of B images and B captions. By computing dense similarity measures
from two directions, we obtain matrices $D_{n,f}$ and $D_{f,n}$ of size $B \times B$, whose diagonal elements correspond to the
image-caption pairs.
Then, we define the agreement loss as: 

\begin{equation}
\label{eq:agree}
\mathcal{L}_{agree} = MSE(diag(D_{n,f}), diag(D_{f,n}))
\end{equation}

where each entry of the similarity matrices $D_{n,f}$ and $D_{f,n}$ results from the computation of
$sim_d(F_k, N_k)$ and $sim_d(N_k, F_k)$ for  \textit{k}th image-caption pair in the batch, respectively.
$diag(\cdot)$ denotes the diagonal elements of a matrix.

\textbf{Total Loss} We combine Equations \ref{eq:contras_fn}, \ref{eq:contras_nf}, and \ref{eq:agree} to obtain the final loss function:

\begin{equation}
\mathcal{L} =  \mathcal{L}_{f,n} + \mathcal{L}_{n,f} + \alpha \mathcal{L}_{agree}
\label{eq:total_loss}
\end{equation}
where $\alpha$ is a hyperparameter.

\subsection{Two-Stage Training with Bootstrapping}
\label{sec:bootstrapping}

We extend SECLA with a two-stage training strategy to adapt it to the need for learning new faces and names from the stream of news in a real-world scenario.
For stage1 training, we train 
a SECLA model (stage1 model) on an easy subset ($D_{easy}$) of the full dataset ($D_{all}$), which contains one or two potential links in each pair. We 
denote the set of names that appear in stage1 training without repetition as $N_{unique}$.
Next, we present the different strategies for stage 2 training in this part.

We first present a two-stage training strategy using simple heuristics in Algorithm \ref{alg:incre_heuristic}. After stage1 training, we split the rest of dataset $D_{rest}$ into two subsets, namely, $D_{match}$ and $D_{unmatch}$, where $D_{match}$ contains all the pairs 
matched by the stage1 model\footnote{The matching is based on inference by computing $sim(f_i, n_j)$ with projected features from $g_{stage1}$ for face $f_i$ and name $n_j$. We match the names in $N_{unique}$ with the faces.}, and $D_{unmatch}$ contains all the unmatched pairs.
Then, we fine-tune the stage1 model on $D_{unmatch}$ with the same loss used for stage1 training.

\begin{algorithm}
\caption{Two-stage
Training using Simple Heuristics}\label{alg:incre_heuristic}
\SetAlgoVlined
\Input{Easy subset  $\mathcal{D}_{easy}$, complementary subset $\mathcal{D}_{rest}$, parameter $\theta_o$ of stage1 model $g_{stage1}(\cdot)$ trained on $\mathcal{D}_{easy}$, Set of unique names $N_{unique}$ in $\mathcal{D}_{easy}$ }
\Output{Stage 2 model $g_{stage2}(\cdot)$}

\textbf{Initialize}:
$\mathcal{D}_{match} = \emptyset$ and contains pairs matched in $\mathcal{D}_{rest}$ using the stage1 model

\For{ Each image-caption pair $\{I, S\} \in \mathcal{D}_{rest}$ contains $\{ f_i, n_j \}, i=1,\dots n, j=1, \dots m $ }
{
\If {$n_j \in N_{unique}$}
  {$f_{aligned} = arg \max_{f_i \in F}(g_{stage1} (f_i, n_j))$ 
  $\mathcal{D}_{match} \leftarrow \mathcal{D}_{match} \cup \{ f_{aligned}, n_j \} $
  }
\Else{Continue}
}
 Let $\mathcal{D}_{unmatch} = \mathcal{D}_{rest} \setminus \mathcal{D}_{match}$ 
 
 $g_{stage2}(\cdot)$ $\leftarrow$ \textit{Fine-tune}($ \mathcal{L}$, $g_{stage1}(\cdot) $, $\mathcal{D}_{unmatch}$)

\end{algorithm}

Using simple heuristics, the trained model can obtain reasonably good performance, as shown below, given a stage1 model with perfect alignment results. However, (1) a stage1 model with 100\% correctness cannot always be expected,
(2) Even with a stage1 model that achieves perfect alignment results for $D_{easy}$, we can still make mistakes while matching new faces in $D_{rest}$ with known names, leading to errors in $D_{unmatch}$ for fine-tuning the model.
(3) As more new names and faces are added during training, the model tends to forget the known names and faces.

To tackle the problems of Algorithm \ref{alg:incre_heuristic}, we propose a two-stage training strategy combined with bootstrapping, as shown in Algorithm \ref{alg:incre}; it includes new loss functions to prevent forgetting and learns better alignments.

\begin{algorithm}
\caption{Two-stage Training with Bootstrapping (SECLA-B)}\label{alg:incre}

\Input{Easy subset  $\mathcal{D}_{easy}$, Full dataset $\mathcal{D}_{all}$, Stage1 model $g_{stage1}(\cdot)$ trained on $\mathcal{D}_{easy}$ with parameter $\theta_o$,  set of unique names and the corresponding faces seen in stage1 $\{ N_{unique}, F_{seen} \} \subset \mathcal{D}_{easy}$ }

\Output{Stage2 model $g_{stage2}(\cdot)$}

\For {Each batch with B image-caption pairs}
{

\textbf{Initialize} 
 batch of matched names and faces $B_{match} = \{\}$,  batch of unmatched names and faces $B_{unmatch} = \{\}$, model $g(\cdot)$ with $\theta_0$ of $g_{stage1}(\cdot)$ 

\For{Each pair $\{I, S\} \in \text{batch of } \mathcal{D}_{all}$, containing $\{ f_i, n_j \}, i=1,\dots n, j=1, \dots m $ }{
\If {$n_j \in N_{unique}$}
   {$f_{aligned} = arg \max_{f_i \in F}(g (n_j, f_i))$
 $B_{match} \leftarrow B_{match}  \cup \{ f_{aligned}, n_j \}$
   }
\Else{$B_{unmatch} \leftarrow B_{unmatch} \cup \{ f_{i}, n_j \}$}

}

\If{$len(B_{match}) > 0$}{$g_{stage2}(\cdot) \leftarrow$ \textit{Train}($\mathcal{L}$, $g(\cdot)$, $B_{unmatch}$) + \textit{Train}($\mathcal{L}_{stage2}$, $g(\cdot)$, $B_{match}$)
}
\Else{ $g_{stage2}(\cdot) \leftarrow$ \textit{Train}($\mathcal{L}$, $g(\cdot)$, $B_{unmatch}$)}
}
\end{algorithm}

After matching the faces using $N_{unique}$ and the stage1 model, we consider the remaining unmatched pairs in 
$B_{unmatch}$ that are not seen in stage1 training. 
Naturally, we can treat these pairs the same way as the pairs in $D_{easy}$ in stage1 training.
Hence, we apply the same loss used for stage1 training, as presented in Equation \ref{eq:total_loss}, for $B_{unmatch}$.

For $B_{match}$, it is desired that our model can maximize the dense similarity measures between the set of matched faces $F'_i$ and the set of matched names $N'_{j}$ from the same sample. Inspired by Li \etal~\cite{li2021prototypical}, we propose a contrastive loss to pull matched faces closer to their respective names in the same sample but push them away from the retrieved prototype faces in other samples. Specifically, given $k$ samples of image-caption pairs with matched name sets $\{N'_1, N'_2, ..., N'_k \}$, matched face sets $\{F'_1, F'_2, ..., F'_k \}$ and retrieved prototype sets $\{ P_1, P_2, ..., P_k \} $, we have:

\begin{equation}
\resizebox{0.9\columnwidth}{!}{%
$
\mathcal{L}_{f,n,p} = -log \frac{e^{sim_d(F'_i, N'_i)}}{e^{sim_d(F'_i, N'_i)} + \sum_{P_j \in P \setminus \{P_i\} } e^{sim_d(P_j, N'_i)} }
$
}
\end{equation}

\begin{equation}
\resizebox{0.9\columnwidth}{!}{%
$
 \mathcal{L}_{n,f,p} = -log \frac{ e^{sim_d(N'_i, F'_i)} }{ e^{sim_d(N'_i, F'_i)}  + \sum_{P_j \in P \setminus \{P_i\} } e^{sim_d(N'_i, P_j)} }
$
}
\end{equation}

We force the symmetry from the two directions as:
\begin{equation}
    \mathcal{L}_{f,n,p}^{B} = \mathcal{L}_{f,n,p} + \mathcal{L}_{n,f,p}
\end{equation}

In addition, to learn more clustered representations of the faces, it is desired that we can pull the representations of matched faces closer to their prototypes in the same sample and push them away from other prototypes. We design a contrastive loss that operates on the face-prototype level as: 

\begin{equation}
\mathcal{L}_{f,p} = -log \frac{ e^{sim_d(F'_i, P_i)} }{\sum_{P_j \in P} e^{sim_d(F'_i, P_j)}}
\end{equation}

\begin{equation}
\mathcal{L}_{p,f} = -log \frac{ e^{sim_d(P_i, F'_i)} }{\sum_{P_j \in P} e^{sim_d(P_j, F'_i)}}
\end{equation}

We denote the bidirectional loss as:
\begin{equation}
    \mathcal{L}_{f,p}^{B} = \mathcal{L}_{f,p} + \mathcal{L}_{p,f}
\end{equation}

Finally, we denote the additional loss for $B_{match}$ as:
\begin{equation}
\mathcal{L}_{stage2} = \mathcal{L}_{f,n,p}^{B} + \mathcal{L}_{f,p}^{B}
\label{eq:stage2}
\end{equation}

In practice, given the known names and faces from stage1 training, different types of prototype faces can be selected: (1) randomly sampled faces, (2) average faces, (3) medoid faces with the minimum average distance to other faces of the same people, and (4) faces matched by the model because they have the highest similarity score with a name.
However, due to factors such as orientation, blurry conditions or lighting conditions, face representations can vary greatly even for the same person. 
Depending on the data distribution, the model performance is affected by different 
types of prototype faces. We discuss the effect of different choices of prototype faces in Sec. \ref{sec:secla-b}.

\subsection{Alignment Scheme}
\label{sec:alignment}

We align faces with names according to the similarity scores from $sim(f_{i}, n_{j}) = (\mathrm{X}_{f_{p}}^i)^\mathrm{T}\cdot \mathrm{X}_{n_{p}}^j$. Specifically, the alignment scheme consists of three parts:

\begin{itemize}
\setlength\itemsep{0.1em}
\item For both training and inference, we add additional NONAME to all the name lists without NONAME.\footnote{Since "NOFACE" links are not included in the Celebrity Together (CelebTo) dataset, we do not add constant NOFACE embeddings to the face list during inference.} Then, NONAME is treated as an extra name. The NONAME embedding is fixed during training.
\item For face $f_i$, if $\text{arg} \max_{m} A_{i,j}$ = $n_m$, we align $f_i$ to $n_m $.
\item If "NOFACE" links are considered, after aligning all the faces, we align the remaining names to NOFACE.
\end{itemize}

The alignment scheme is applied during inference.

\section{Experimental Setup}

\begin{table*}[t]
\begin{center}
\begin{tabular}{ccccccc}
  \hline
  \multirow{2}{*}{Model} & \multirow{2}{*}{Contrastive} & \multirow{2}{*}{Symmetry}  & \multicolumn{3}{c}{LFW}
  & CelebTo \\
    & &  & Precision & Recall & F1 & Accuracy \\
  \hline
 Pham \etal~\cite{5332299} (using $P(f|n)$ ) & & & 74.90\%
 & 70.56\% & 72.66\%  & \slash \\
  Pham \etal~\cite{5332299} (using $P(n|f)$ ) & & & 69.99\% &72.73\% & 71.33\% & \slash \\
  Wang \etal~\cite{conf/emnlp/WangTSMY20} ($\mathcal{L}_{f,n}$) & \checkmark & & 41.52\%  & 45.92\% & 43.61\%  &  44.77\% \\
  Wang \etal~\cite{conf/emnlp/WangTSMY20} ($\mathcal{L}_{n,f} $) & \checkmark & & 37.46\% & 41.42\% & 39.34\% & 41.29\%   \\
 Hesssel \etal~\cite{hessel-etal-2019-unsupervised} (w/o TK) & & \checkmark & 61.35\% & 67.84\% & 64.43\% & 45.70\% \\
  Hesssel \etal~\cite{hessel-etal-2019-unsupervised} (TK) & & \checkmark & 63.51\% & 70.23\% & 66.70\% & 40.31\% \\
  \hline
  SECLA-$\alpha \mathcal{L}_{agree}$ (ours) & \checkmark & \checkmark  & 74.40\% &  82.28\% & 78.14\% & 81.79\% \\
  SECLA (ours) & \checkmark & \checkmark  & 76.96\% & 85.11\% & 80.83\%  & 87.46\%  \\
  SECLA-B (ours) & \checkmark & \checkmark & \textbf{77.94\%} & \textbf{86.19\%} & \textbf{81.86\%}  & \textbf{88.36\%}  \\
  \hline
\end{tabular}
\end{center}
\caption{Performance Comparison of the Alignment Models on the LFW and CelebTo Datasets. We compare our results with the best results from \cite{5332299}, the results of the model trained with the loss function in \cite{conf/emnlp/WangTSMY20} (which is equivalent to our single-directional loss function), and the results of the model trained using the dense similarity measure with hinge loss and the top k selection strategy (TK), as in \cite{hessel-etal-2019-unsupervised}.} \label{tab:overall}
\end{table*}

\subsection{Datasets}

We conduct extensive experiments on the same LFW dataset used in \cite{5332299}. Following \cite{5332299}, we extract names with the named entity recognizer from OpenNLP package\footnote{http://opennlp.sourceforge.net/} and detect bounding boxes for faces with the OpenCV implementation of \cite{Viola04a}. We obtain 10976 pairs of news images and captions, covering 23928 links, from LFW. On average, 1.97 names and 1.32 faces appear in each pair. 

We also use the Celebrity Together (CelebTo) dataset
\cite{conf/eccv/ZhongAZ18}.
CelebTo has the same numbers of faces and names in each pair.
It does not include "NOFACE" links 
and no captions are provided.
It contains 193523 images with 545519 faces and 2533 identities. The average number of faces in each image is 2.82, 
and that of NONAME is 1.16.

\subsection{Implementation Details}

We adopt Inception ResNet (V1) \cite{Szegedy:2017vx} pre-trained on VGGFace2 \cite{conf/fgr/CaoSXPZ18} as the face feature extractor and BERT-based \cite{devlin-etal-2019-bert} as the name feature extractor.
The name representation is the average embedding from all the tokens of the summation of the last four layers of BERT.
We fix the embedding for "NONAME"
as that for the "[UNK]" token.
We construct a one-layer MLP as the name projector $g_{n}(\cdot): \mathbb{R}^{768} \rightarrow  \mathbb{R}^{512}$ and a three-layer MLP as the common projector $g_{c}(\cdot): \mathbb{R}^{512} \rightarrow  \mathbb{R}^{128}$. We use ReLU as the activation function.
We set the hyperparameter $\alpha$=$0.15$ in Equation \ref{eq:total_loss} and the threshold in the hinge loss to 0.2 \cite{hessel-etal-2019-unsupervised}.
We train our model using the Adam optimizer \cite{kingma2015method} with a learning rate=$3\times 10^{-4}$.
We train SECLA with a batch size of 20 for 30 epochs and 3 epochs for LFW and CelebTo, respectively.
For SECLA-B, the stage1 model is trained on the subset ($D_{one}$) with one name and one face and the subset ($D_{2name}$) with two names and two faces for LFW and CelebTo, respectively. We exclude NONAME from both sets.
Stage1 training takes 15 epochs and 5 epochs, and stage2 training takes 20 epochs and 2 epochs for LFW and CelebTo, respectively.
The models are trained on one NVIDIA TITAN Xp GPU.

\subsection{Evaluation Metrics}

Due to the existence of null links in the LFW dataset that might refer to different names or faces, 
we do not use accuracy \cite{NIPS2004_03fa2f75, 1315253} to evaluate aligned links. Following \cite{5332299}, we adopt precision, recall, and F1 score as the evaluation metrics, where
\begin{equation}
\resizebox{0.885\columnwidth}{!}{
    $\text{Precision} = \frac{\# \text{link}_{\text{correct}}}{ \# \text{link}_{\text{found}} }$, $\text{Recall} = \frac{\# \text{link}_{\text{correct}}}{ \# \text{link}_{\text{gt}} }$, 
    $\text{F1} = \frac{  2\text{Recall} \cdot \text{Precision} }{ \text{Recall}+\text{Precision} }$
}
\label{eq:evaluation}
\end{equation}
For CelebTo, we use accuracy for evaluation.

\section{Results}
\label{sec:result}

\subsection{Overall Performance}

In Table \ref{tab:overall}, we compare different alignment models on the LFW and CelebTo datasets. For LFW, we choose the best performance reported in \cite{5332299} in terms of F1 score and recall for comparison. We also implement models trained with the loss proposed in \cite{conf/emnlp/WangTSMY20}, which is equivalent to the unidirectional version of our contrastive loss, as well as the learning strategy proposed in \cite{hessel-etal-2019-unsupervised}. For CelebTo, we consider the face-name alignment task as a metric learning problem by excluding all the context information (e.g., syntactic structures of the captions, co-occurrence of faces in images) from training, which was heavily relied on in \cite{5332299}.

Our SECLA model outperforms the previous SOTA and other approaches from related fields, as shown in Table \ref{tab:overall}.
In particular, we find that for LFW, SECLA improves the results of the model in \cite{5332299} by a large margin in terms of recall and F1 but only achieves slightly better precision (74.90\% vs 76.96\%). This is because SECLA always identifies all the links within an image-caption pair, in contrast to \cite{5332299}, which only finds parts of the links above a certain threshold. According to Equation \ref{eq:evaluation}, we calculate precision as the number of correct links divided by the number of links found, so it is expected that we cannot achieve high precision in that case.
Moreover, the models trained with unidirectional loss $\mathcal{L}_{f,n}$ and $\mathcal{L}_{n,f}$ achieve very poor performance with F1 scores of 43.61\% and 39.34\%, respectively, on LFW and accuracy of 44.77\% and 41.29\%, respectively, on CelebTo.
We randomly initialize our projectors so that the semantic meaning of face features and name features are not close enough, leading to the initialized scores for different names and faces falling into a small interval near zero (mostly [-0.1, 0.1]).
In that case, without proper symmetry-enhanced regularization, instead of optimizing the similarity scores for matched pairs toward the positive side, the model randomly optimizes the scores. The model in \cite{hessel-etal-2019-unsupervised} achieves relatively better results on LFW but performs much worse on CelebTo. With the strong constraint of margins in the decision boundary, the hinge loss used in \cite{hessel-etal-2019-unsupervised} makes the lack of semantic similarities of faces and names at the early stage of training a big problem. Our contrastive loss functions in Equations \ref{eq:contras_fn} and \ref{eq:contras_nf} do not have this problem and thus achieve significantly better results.

Adding agreement loss increases the performance of SECLA from F1=78.14\% to 80.83\% on LFW, and accuracy=81.79\% to 87.46\% on CelebTo. The agreement loss works better when more faces and names appear.

\begin{table}[!htbp]
\begin{tabular}{ccccc}
  \hline
  \multirow{2}{*}{Model}  & \multicolumn{3}{c}{LFW}
  & CelebTo \\
      & Precision & Recall & F1 & Accuracy \\
  \hline
  Pipeline  & 69.94\% & 77.34\% & 73.46\%  & 67.50\%  \\
  SECLA  & 76.96\% & 85.11\% & 80.83\%  & 87.46\%  \\
  
  \hline
 
  SECLA-B & \textbf{77.94\% } & \textbf{86.19\%} & \textbf{81.86\%} & \textbf{88.36\%}   \\
  \hline
\end{tabular}
\caption{Performance comparison of SECLA-B on LFW and CelebTo. The Pipeline model is trained with Algorithm \ref{alg:incre_heuristic}.}\label{tab:bootstrapping}
\end{table}

We present the performance comparison of SECLA-B with the baselines in Table \ref{tab:bootstrapping}.
We take SECLA and the model learned with Algorithm \ref{alg:incre_heuristic} (denoted as Pipeline) as baselines.
As shown in Table \ref{tab:bootstrapping}, 
we achieve good results with Pipeline; i.e., F1 = $73.46\%$ on LFW and accuracy = $67.50\%$ on CelebTo.
However, the performance of Pipeline is much worse than that of SECLA.
Based on Algorithm \ref{alg:incre_heuristic}, the errors made by the stage1 model accumulate during fine-tuning.
Specifically, after stage1 training on an easy subset, 
we split the rest of the dataset 
into $D_{match}$ and $D_{unmatch}$ and fine-tune the stage1 model on $D_{unmatch}$. $D_{match}$ contains pairs of matched faces and names, which are not guaranteed to be correctly aligned.
Even if SECLA achieves perfect alignment results for stage1 training, for a new face of a known name, the model could still make mistakes. Incorrect matches made by the stage1 model result in incorrectly matched faces and names in $D_{unmatch}$, which is used for fine-tuning the stage1 model. Hence, the error accumulates and causes performance degradation. 
Despite
 the errors made by the stage1 model, SECLA-B further improves the performance of the strong SECLA model.

\subsection{Case Study for SECLA}

\noindent \textbf{Experiments on $D_{easy}$}

In this part, we apply the SECLA model on easy subset $D_{easy}$, which is also the stage1 training of SECLA-B. The performance of the stage1 model is vital for SECLA-B since we have to rely on it for matching faces in stage2 training.

For LFW, we extract a subset with one face in each pair ($D_{one}$) as $D_{easy}$.
We exclude NONAME and NOFACE to decrease the level of ambiguity.
The best alignment results achieved by \cite{5332299} are precision = $85.41\%$, recall = $91.17\%$ and F1 = $88.19\%$. Aligning one name to one face is our task in its easiest form, yet the model 
in \cite{5332299} confuses common names with NONAME in approximately 10\% of the cases. For comparison, our SECLA model achieves perfect alignment results by correctly aligning all the links.

In CelebTo, 
each image contains at least two faces.
Thus, we extract a subset with two faces in each pair ($D_{2name}$) as $D_{easy}$;
it contains 53334 samples with 2437 unique names.
Although this is a more difficult task, where we have to align two names to two faces in each sample, the majority of the names occur more than 10 times in $D_{2name}$, making the task slightly easier.

\begin{table}
\centering
\begin{tabular}{lcc}
  \hline
   Model  & Add NONAME  & Accuracy \\
  \hline
  SECLA  & \checkmark & \textbf{96.26\%}  \\
  SECLA  &  & 90.52\%  \\
  SECLA-$\alpha \mathcal{L}_{agree}$  & \checkmark  & 95.27\%  \\
  
  \hline
\end{tabular}
\caption{Performance of SECLA on $D_{2name}$ subset of CelebTo.} \label{tab:2name}
\end{table}


In addition, we conduct ablations of SECLA on $D_{2name}$.
As presented in Table \ref{tab:2name}, without agreement loss, the accuracy decreases from $96.26\%$ to $95.27\%$. Since most names occur frequently in $D_{2name}$, which is favorable for the contrastive loss to differentiate between pairs, the symmetry constraint imposed by agreement loss indeed helps the model to learn more accurate alignments. Adding NONAME in the name list during training, we observe a large improvement in accuracy from $90.52\%$ to $96.26\%$.

\noindent \textbf{Experiments with More Ambiguities}

For cases with more ambiguities, we conduct experiments on a subset of LFW with at least 2 faces. This subset contains 3029 image-caption pairs with 8960 links. Although the number of links is not very high compared to that of image-caption pairs, there are 4177 unique names
within the 8960 links. Most of the names 
only appear once or twice.
Furthermore, we obtain 776 incorrect links, which can mislead the model to learn incorrect alignments. 

We present the results in Table \ref{tab:2face}. The model in \cite{5332299} struggles in this more ambiguous task, i.e., F1 = $47.48\%$. The performance of the unidirectional model varies greatly (F1=$65.45\%$ for name-to-face and F1=$47.87\%$ for face-to-name). With and without agreement loss, our bidirectional model achieves very positive results, i.e., F1=$66.89\%$ and F1=$65.97\%$, respectively. The challenge of this task is that there are too many unique names compared to the number of links, which is not beneficial for the contrastive loss. However, our model still outperforms the previous SOTA \cite{5332299} by a large margin,
and the agreement loss is proven to be useful.
We also observe improvements in performance by adding the agreement loss to unidirectional models.

\begin{table}[!htbp]
\begin{center}
\resizebox{\columnwidth}{!}{
\begin{tabular}{lccc}
  \hline
 Model    & Precision & Recall & F1  \\
    \hline
  Pham \etal~\cite{5332299} & 44.05\% & 51.51\% & 47.48\% \\
   SECLA  &  \textbf{64.00\%} & \textbf{70.06\%} & \textbf{66.89\%} \\
   SECLA -  $\alpha \mathcal{L}_{agree}$ & 63.11\% &  69.10\% & 65.97\%   \\
   SECLA - $\mathcal{L}_{n,f}$ & 61.80\% & 67.66\% & 64.59\% \\
   SECLA - $\mathcal{L}_{f,n}$ & 63.06\% & 69.04\% & 65.91\% \\
   SECLA - $\alpha \mathcal{L}_{agree}$ - $\mathcal{L}_{n,f}$ & 45.79\% & 50.13\% &  47.87\% \\
   SECLA - $\alpha \mathcal{L}_{agree}$ - $\mathcal{L}_{f,n}$ & 62.61\% & 68.55\% & 65.45\% \\
  \hline
\end{tabular}%
}
\end{center}
\caption{Performance of SECLA on LFW with at least 2 faces.} \label{tab:2face}
\end{table}

\subsection{Case Study for SECLA-B}
\label{sec:secla-b}

\noindent \textbf{Choices of Prototype Faces}

As illustrated in Sec. \ref{sec:bootstrapping}, we consider four types of prototype faces, namely, 
(1) a randomly selected face (random face), (2) an average face (avg. face), (3) a face with a minimum average distance to the other faces (medoid face), and (4) a face with the highest similarity score to the respective name (matched face).
We present the best performance achieved by SECLA-B with different choices of prototype faces in Table \ref{tab:proto}. 
For LFW, different choices of prototype faces have limited effects on the performance of SECLA-B. The main reason for that is that most of the names in the subset ($D_{one}$) used for stage1 training occur only once or twice. In that case, the effect of different choices of prototypes is limited. For CelebTo, we observe that using the average face causes significant degradation in performance compared to using other types of prototype faces or even SECLA ($87.46\%$ accuracy on CelebTo). The cause of this is also related to the data distribution. As stated before, only 2437 unique names can be found in 53334 samples of $D_{2name}$. For most identities, due to variations in faces, the average face 
calculated from hundreds of faces is no longer a good representative. In general, CelebTo favors prototype faces with more diversity (random face/matched face).

\begin{table}[!htbp]
\begin{center}
\resizebox{\columnwidth}{!}{%
\begin{tabular}{ccccc}
  \hline
  \multirow{2}{*}{Prototype Type}  & \multicolumn{3}{c}{LFW}
  & CelebTo \\
        & Precision & Recall & F1 & Accuracy \\
  \hline
  random face & 77.65\% & 85.87\% & 81.55\% & \textbf{88.36\%} \\
  avg. face & \textbf{77.94\%} & \textbf{86.19\%} & \textbf{81.86\%} & 81.19\% \\
  medoid face & 77.87\% & 86.12\% & 81.79\% & 86.07\% \\
  matched face & 77.74\% & 85.97\% & 81.61\% & 87.94\% \\
  \hline
\end{tabular}
}
\end{center}
\caption{Best performance achieved by SECLA-B with different choices of prototype faces for stage 2 training.} \label{tab:proto}
\end{table}

\noindent \textbf{Ablation Studies of SECLA-B}

In this part, we present ablation studies of SECLA-B. We show the results of SECLA-B trained with matched faces, and these results are satisfactory for both datasets. We also explore the effect of adding 
the NOFACE to the matched face list, which is not considered for SECLA. Unlike in SECLA, where no matching is realized, incorrectly matched faces could lead to errors that accumulate during stage2 training of SECLA-B.\footnote{Our experiments also show that the additional NOFACE have little effect on the performance of SECLA. We fix the embedding of NOFACE as $[F_{n,1}, F_{n,2}, \ldots F_{n,512}]$ with $F_{n,i}\sim N(0,1)$. }
The additional NOFACE could potentially mitigate this problem.
The results in Table \ref{tab:secla-b_ablation} show that
adding NOFACE is effective, especially for CelebTo, where the stage1 model only correctly aligns 96.26\% of the links. Overall, we achieve the best performance with both $\mathcal{L}_{n,f,p}^{B}$ and $\mathcal{L}_{f,p}^{B}$ used during stage2 training. We observe that $\mathcal{L}_{n,f,p}^{B}$ contributes the most to the SECLA-B performance. And $\mathcal{L}_{f,p}^{B}$ alone is not enough to achieve satisfactory results when the performance of the stage1 model is limited.

\begin{table}[!htbp]
\begin{center}
\resizebox{\columnwidth}{!}{%
\begin{tabular}{lccccc}
  \hline
  \multirow{2}{*}{Model} & \multirow{2}{*}{NOFACE} & \multicolumn{3}{c}{LFW}
  & CelebTo \\
      &  & Precision & Recall & F1 & Accuracy \\
  \hline
  SECLA-B & \checkmark & \textbf{77.74\%} & \textbf{85.97\%} & \textbf{81.61\%} & \textbf{87.94\%} \\
  SECLA-B & & 77.25\% & 85.43\% & 81.13\% & 86.74\% \\
  SECLA-B-$\mathcal{L}_{n,f,p}^{B}$ & \checkmark & 77.47\% & 85.67\% & 81.37\% & 80.90\% \\
  SECLA-B-$\mathcal{L}_{f,p}^{B}$ & \checkmark & 77.62\% & 85.84\% & 81.52\% & 87.11\% \\
  \hline
\end{tabular}%
}
\end{center}
\caption{Ablation study of SECLA-B with matched faces. } \label{tab:secla-b_ablation}
\end{table}

\section{Conclusion}

We present a bidirectional contrastive learning framework (SECLA) enhanced with forced symmetry to address the weakly supervised face-name alignment problem without any annotated training examples or external data mined from the web. The method shows satisfactory performance on LFW and CelebTo. Moreover, we present the SECLA-B model trained with a two-stage learning strategy and bootstrapping; it further improves the performance of SECLA. Our methods can be adapted to other multimodal news understanding tasks, including multimodal summarization and news image captioning, which we leave for future research.

{\small
\bibliographystyle{ieee_fullname}
\bibliography{egpaper}

\begin{thebibliography}{10}\itemsep=-1pt

\bibitem{NIPS2004_03fa2f75}
Tamara Berg, Alexander Berg, Jaety Edwards, and David Forsyth.
\newblock Who\textquotesingle s in the picture.
\newblock In L. Saul, Y. Weiss, and L. Bottou, editors, {\em Advances in Neural
  Information Processing Systems}, volume~17. MIT Press, 2005.

\bibitem{1315253}
T.L. Berg, A.C. Berg, J. Edwards, M. Maire, R. White, Yee-Whye Teh, E.
  Learned-Miller, and D.A. Forsyth.
\newblock Names and faces in the news.
\newblock In {\em Proceedings of the 2004 IEEE Computer Society Conference on
  Computer Vision and Pattern Recognition, 2004. CVPR 2004.}, volume~2, pages
  II--II, 2004.

\bibitem{conf/fgr/CaoSXPZ18}
Qiong Cao, Li Shen, Weidi Xie, Omkar~M. Parkhi, and Andrew Zisserman.
\newblock Vggface2: A dataset for recognising faces across pose and age.
\newblock In {\em FG}, pages 67--74. IEEE Computer Society, 2018.

\bibitem{chen-etal-2021-mask}
Chi Chen, Maosong Sun, and Yang Liu.
\newblock Mask-align: Self-supervised neural word alignment.
\newblock In {\em Proceedings of the 59th Annual Meeting of the Association for
  Computational Linguistics and the 11th International Joint Conference on
  Natural Language Processing (Volume 1: Long Papers)}, pages 4781--4791,
  Online, Aug. 2021. Association for Computational Linguistics.

\bibitem{conf/mir/ChenFNJH15}
Zhineng Chen, Bailan Feng, Chong-Wah Ngo, Caiyan Jia, and Xiangsheng Huang.
\newblock Improving automatic name-face association using celebrity images on
  the web.
\newblock In Alexander~G. Hauptmann, Chong-Wah Ngo, Xiangyang Xue, Yu-Gang
  Jiang, Cees Snoek, and Nuno Vasconcelos, editors, {\em ICMR}, pages 623--626.
  ACM, 2015.

\bibitem{journals/mms/ChenZDXG19}
Zhineng Chen, Wei Zhang, Bin Deng, Hongtao Xie, and Xiaoyan Gu.
\newblock Name-face association with web facial image supervision.
\newblock {\em Multim. Syst.}, 25(1):1--20, 2019.

\bibitem{Cui_2021_ICCV}
Yuqing Cui, Apoorv Khandelwal, Yoav Artzi, Noah Snavely, and Hadar
  Averbuch-Elor.
\newblock Who's waldo? linking people across text and images.
\newblock In {\em Proceedings of the IEEE/CVF International Conference on
  Computer Vision (ICCV)}, pages 1374--1384, October 2021.

\bibitem{devlin-etal-2019-bert}
Jacob Devlin, Ming-Wei Chang, Kenton Lee, and Kristina Toutanova.
\newblock {BERT}: Pre-training of deep bidirectional transformers for language
  understanding.
\newblock In {\em Proceedings of the 2019 Conference of the North {A}merican
  Chapter of the Association for Computational Linguistics: Human Language
  Technologies, Volume 1 (Long and Short Papers)}, pages 4171--4186,
  Minneapolis, Minnesota, June 2019. Association for Computational Linguistics.

\bibitem{conf/cvpr/FangGISDDGHMPZZ15}
Hao Fang, Saurabh Gupta, Forrest~N. Iandola, Rupesh~Kumar Srivastava, Li Deng,
  Piotr Dollár, Jianfeng Gao, Xiaodong He, Margaret Mitchell, John~C. Platt,
  C.~Lawrence Zitnick, and Geoffrey Zweig.
\newblock From captions to visual concepts and back.
\newblock In {\em CVPR}, pages 1473--1482. IEEE Computer Society, 2015.

\bibitem{conf/eccv/FarhadiHSYRHF10}
Ali Farhadi, Seyyed Mohammad~Mohsen Hejrati, Mohammad~Amin Sadeghi, Peter
  Young, Cyrus Rashtchian, Julia Hockenmaier, and David~A. Forsyth.
\newblock Every picture tells a story: Generating sentences from images.
\newblock In Kostas Daniilidis, Petros Maragos, and Nikos Paragios, editors,
  {\em ECCV (4)}, volume 6314 of {\em Lecture Notes in Computer Science}, pages
  15--29. Springer, 2010.

\bibitem{conf/cvpr/GuillauminMVS08}
Matthieu Guillaumin, Thomas Mensink, Jakob~J. Verbeek, and Cordelia Schmid.
\newblock Automatic face naming with caption-based supervision.
\newblock In {\em CVPR}. IEEE Computer Society, 2008.

\bibitem{hessel-etal-2019-unsupervised}
Jack Hessel, Lillian Lee, and David Mimno.
\newblock Unsupervised discovery of multimodal links in multi-image,
  multi-sentence documents.
\newblock In {\em Proceedings of the 2019 Conference on Empirical Methods in
  Natural Language Processing and the 9th International Joint Conference on
  Natural Language Processing (EMNLP-IJCNLP)}, pages 2034--2045, Hong Kong,
  China, Nov. 2019. Association for Computational Linguistics.

\bibitem{conf/nips/KarpathyJL14}
Andrej Karpathy, Armand Joulin, and Fei-Fei Li.
\newblock Deep fragment embeddings for bidirectional image sentence mapping.
\newblock In Zoubin Ghahramani, Max Welling, Corinna Cortes, Neil~D. Lawrence,
  and Kilian~Q. Weinberger, editors, {\em NIPS}, pages 1889--1897, 2014.

\bibitem{kingma2015method}
Diederik~P. Kingma and Jimmy Ba.
\newblock Adam: A method for stochastic optimization.
\newblock In Yoshua Bengio and Yann LeCun, editors, {\em ICLR (Poster)}, 2015.

\bibitem{li2021prototypical}
Junnan Li, Pan Zhou, Caiming Xiong, and Steven Hoi.
\newblock Prototypical contrastive learning of unsupervised representations.
\newblock In {\em International Conference on Learning Representations}, 2021.

\bibitem{liang-etal-2006-alignment}
Percy Liang, Ben Taskar, and Dan Klein.
\newblock Alignment by agreement.
\newblock In {\em Proceedings of the Human Language Technology Conference of
  the {NAACL}, Main Conference}, pages 104--111, New York City, USA, June 2006.
  Association for Computational Linguistics.

\bibitem{5332299}
Phi~The Pham, Marie-Francine Moens, and Tinne Tuytelaars.
\newblock Cross-media alignment of names and faces.
\newblock {\em IEEE Transactions on Multimedia}, 12(1):13--27, 2010.

\bibitem{schroff2015facenet}
Florian Schroff, Dmitry Kalenichenko, and James Philbin.
\newblock Facenet: A unified embedding for face recognition and clustering.
\newblock In {\em Proceedings of the IEEE conference on computer vision and
  pattern recognition}, pages 815--823, 2015.

\bibitem{journals/mta/SuPFW16}
Xueping Su, Jinye Peng, Xiaoyi Feng, and Jun Wu.
\newblock Labeling faces with names based on the name semantic network.
\newblock {\em Multimedia Tools Appl.}, 75(11):6445--6462, 2016.

\bibitem{journals/mta/SuPFWFC14}
Xueping Su, Jinye Peng, Xiaoyi Feng, Jun Wu, Jianping Fan, and Li Cui.
\newblock Cross-modality based celebrity face naming for news image
  collections.
\newblock {\em Multimedia Tools Appl.}, 73(3):1643--1661, 2014.

\bibitem{Szegedy:2017vx}
Christian Szegedy, Sergey Ioffe, Vincent Vanhoucke, and Alexander~A Alemi.
\newblock {Inception-v4, Inception-ResNet and the Impact of Residual
  Connections on Learning}.
\newblock In {\em Proceedings of the 31st AAAI Conference on Artificial
  Intelligence}, AAAI '17, pages 4278--4284. AAAI Press, 2017.

\bibitem{8880535}
Yong Tian, Lian Zhou, Yuejie Zhang, Tao Zhang, and Weiguo Fan.
\newblock Deep cross-modal face naming for people news retrieval.
\newblock {\em IEEE Transactions on Knowledge and Data Engineering},
  33(5):1891--1905, 2021.

\bibitem{conf/cvpr/VinyalsTBE15}
Oriol Vinyals, Alexander Toshev, Samy Bengio, and Dumitru Erhan.
\newblock Show and tell: A neural image caption generator.
\newblock In {\em CVPR}, pages 3156--3164. IEEE Computer Society, 2015.

\bibitem{Viola04a}
Paul Viola and Michael Jones.
\newblock Robust real-time object detection.
\newblock {\em International Journal of Computer Vision}, 57(2):137--154, 2004.

\bibitem{conf/emnlp/WangTSMY20}
Qinxin Wang, Hao Tan, Sheng Shen, Michael~W. Mahoney, and Zhewei Yao.
\newblock Maf: Multimodal alignment framework for weakly-supervised phrase
  grounding.
\newblock In Bonnie Webber, Trevor Cohn, Yulan He, and Yang Liu, editors, {\em
  EMNLP (1)}, pages 2030--2038. Association for Computational Linguistics,
  2020.

\bibitem{conf/icml/XuBKCCSZB15}
Kelvin Xu, Jimmy Ba, Ryan Kiros, Kyunghyun Cho, Aaron~C. Courville, Ruslan
  Salakhutdinov, Richard~S. Zemel, and Yoshua Bengio.
\newblock Show, attend and tell: Neural image caption generation with visual
  attention.
\newblock In Francis~R. Bach and David~M. Blei, editors, {\em ICML}, volume~37
  of {\em JMLR Workshop and Conference Proceedings}, pages 2048--2057.
  JMLR.org, 2015.

\bibitem{conf/ijcai/ZhangWLJXF13}
Yuejie Zhang, Wei Wu, Yang Li, Cheng Jin, Xiangyang Xue, and Jianping Fan.
\newblock Automatic name-face alignment to enable cross-media news retrieval.
\newblock In Francesca Rossi, editor, {\em IJCAI}, pages 2768--2775.
  IJCAI/AAAI, 2013.

\bibitem{conf/eccv/ZhongAZ18}
Yujie Zhong, Relja Arandjelovic, and Andrew Zisserman.
\newblock Compact deep aggregation for set retrieval.
\newblock In Laura Leal-Taixé and Stefan Roth, editors, {\em ECCV Workshops
  (4)}, volume 11132 of {\em Lecture Notes in Computer Science}, pages
  413--430. Springer, 2018.

\end{thebibliography}
}

\end{document}